\documentclass[runningheads]{llncs}

\usepackage{eccv}

\usepackage{eccvabbrv}

\usepackage{graphicx}
\usepackage{booktabs}

\usepackage{amsmath} 
\usepackage{amsfonts} 
\usepackage{amssymb} 
\usepackage{bm}
\usepackage{multirow}
\usepackage{makecell}

\usepackage{xcolor}
 
\newcommand{\x}{{\bf x}}

\usepackage[accsupp]{axessibility}  

\usepackage{hyperref}

\usepackage{orcidlink}

\begin{document}

\title{Generalizable Facial Expression Recognition} 

\titlerunning{Generalizable FER}

\author{Yuhang Zhang\orcidlink{0000-0003-4161-5020} \and
Xiuqi Zheng \and
Chenyi Liang \and
Jiani Hu \and Weihong Deng}

\authorrunning{Y. Zhang et al.}

\institute{Beijing University of Posts and Telecommunications\\
\email{\{zyhzyh, xiuqizheng, liangchenyi, jnhu, whdeng\}@bupt.edu.cn}}

\maketitle

\begin{abstract}
  SOTA facial expression recognition (FER) methods fail on test sets that have domain gaps with the train set. Recent domain adaptation FER methods need to acquire labeled or unlabeled samples of target domains to fine-tune the FER model, which might be infeasible in real-world deployment. \emph{In this paper, we aim to improve the zero-shot generalization ability of FER methods on different unseen test sets using \textbf{only one train set}.} Inspired by how humans first detect faces and then select expression features, we propose a novel FER pipeline to extract expression-related features from any given face images. Our method is based on the generalizable face features extracted by large models like CLIP. However, it is non-trivial to adapt the general features of CLIP for specific tasks like FER. To preserve the generalization ability of CLIP and the high precision of the FER model, we design a novel approach that learns sigmoid masks based on the fixed CLIP face features to extract expression features. To further improve the generalization ability on unseen test sets, we separate the channels of the learned masked features according to the expression classes to directly generate logits and avoid using the FC layer to reduce overfitting. We also introduce a channel-diverse loss to make the learned masks separated. Extensive experiments on five different FER datasets verify that our method outperforms SOTA FER methods by large margins. Code is available in \url{https://github.com/zyh-uaiaaaa/Generalizable-FER}.
  \keywords{Facial expression recognition \and Zero-shot generalization \and Mask learning}
\end{abstract}

\section{Introduction}
\label{sec:intro}

\begin{figure}[t]
  \centering
  \includegraphics[width=1.\linewidth]{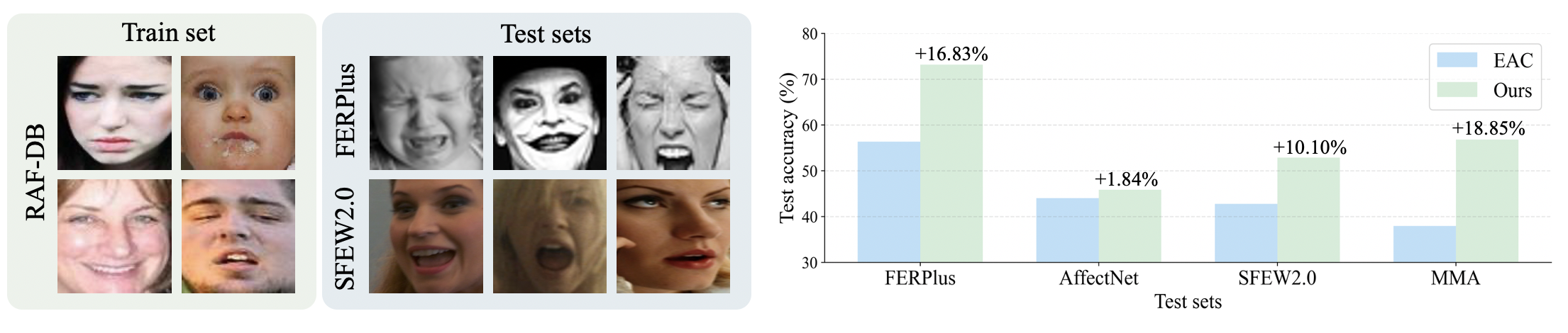}
    \caption{The illustration of Generalizable Facial Expression Recognition (GFER). In order to evaluate the generalization ability of FER methods, we train FER models on one train set and test the trained models on different unseen test sets. \emph{The difference between our task and domain adaptation FER is that we only use \textbf{one} train set and \textbf{do not} acquire any labeled or unlabeled samples from the target domain.} SOTA FER methods like EAC do not work well on unseen test sets, which show low generalization ability. Our method 
    outperforms EAC by large margins on different unseen FER test sets, showing better generalization ability.}
    \label{figure1}
\end{figure}

Facial expression recognition (FER) aims to understand human feelings and is vital to human-computer interaction. With the development of deep learning, FER methods achieve great success on both laboratory-collected and in-the-wild FER datasets. However, we find that things are different when the test sets have domain gaps with the train set. 
For example, if we train the SOTA FER model EAC~\cite{zhang2022learn} on one given FER dataset like RAF-DB~\cite{li2017reliable}, then it can only achieve high performance on the test set of RAF-DB, while very low performance on other different FER test sets like AffectNet~\cite{mollahosseini2017affectnet}, shown in Fig.~\ref{figure1}. Though some works try to deal with the domain generalization problem in FER, they all assume accessing labeled or unlabeled target set samples for FER model fine-tuning. However, in the real-world FER, as we do not previously know the distribution of the test samples, we might not be able to get access to even the unlabeled target samples. In such cases, the domain adaptation FER methods cannot work. \emph{In this paper, we aim to improve the zero-shot generalization ability of FER methods on different unseen test sets using only one train set.} The difference between our work and existing domain adaptation FER works is that we only utilize \textbf{one} given train set and we \textbf{do not} utilized labeled or unlabeled target samples to fine-tune our model. This is more similar to the real-world deployment where we usually only have one train set with a similar style of annotating (\emph{without considering annotation inconsistency}) and we need to recognize unseen test samples from various domains. As the generalization ability of the FER model is our main focus, we name the paper as Generalizable Facial Expression Recognition (GFER).

We find in our experiments that SOTA FER methods achieve low test accuracy on different unseen FER test sets. The reason might lie in that these methods fit the given FER train set too well (including domain-specific information) to predict on test sets with domain gaps. However, we humans take a very generalizable way to recognize facial expressions. Given images with domain gaps, we first locate the face features, then we disentangle expression features from the face features and make a recognition only based on the expression features. Inspired by that, we design a novel pipeline to improve the generalization ability of FER methods. We first extract generalizable face features from pre-trained large models like CLIP. As the large models have been trained with numerous images including face images, we can assume the extracted face features from them could generalize seamlessly to different FER domains. However, it is non-trivial to adapt the general features extracted by CLIP for specific tasks like FER. We further propose a novel method to learn sigmoid masks to select expression-related features for FER. Specifically, we fix the general face features during the whole training process to maintain the generalization ability. 
We then train a FER model to learn masks based on these general face features. 
In order to make the learned masks generalizable instead of overfitting the given face features, we propose to utilize a sigmoid function to regularize the learned masks. The sigmoid mask is vital to the success of method, which is discussed in our experiments. Furthermore, enlightened by the channel-wise attention and the fixed number of the seven basic expressions, we separate the masks to respond to different expressions according to the channels. This operation avoids using the FC layer, which further reduces the overfitting probability of the FER model. Finally, we introduce a channel-diverse loss to make the masks corresponding to different expressions as diverse as possible to improve the generalization ability of the learned masks.  

Extensive experiments on five different FER datasets validate the effectiveness of our proposed method. We carry out ablation study and design comparison groups to study each part of our method. Visualization results of learned features and sigmoid masks are provided to help understanding. We summarize our main contributions as follows.

\begin{itemize}
\item We propose to learn sigmoid masks on fixed general face features for FER. This novel FER pipeline improves zero-shot generalization ability on different unseen test sets using only one train set. 
\item  We design a channel-separation module and a channel-diverse loss to further improve the generalization ability of the learned sigmoid masks. Channel-separation modules separate the learned masked features according to the channel dimension to reduce overfitting. The channel-diverse loss regularizes the sigmoid masks to be as diverse as possible.
\item Extensive experiments on five different FER methods illustrate that our method outperforms SOTA FER methods by remarkable margins on different unseen test sets, which shows the superior generalization ability of our proposed method.
\end{itemize}

\section{Related Work}
\label{sec:formatting}
\subsection{Facial Expression Recognition} 
Facial Expression Recognition (FER) plays a vital role in human-computer interaction, and extensive research has been conducted to enhance the performance of FER~\cite{shan2009facial, zhi2010graph, zhong2012learning, bargal2016emotion, kahou2013combining, farzaneh2021facial, ruan2021feature, li2017reliable, li2023unconstrained, li2022crs, li2021lban, li2021adaptively, zeng2022face2exp, jin2021learning, kim2022emotion}. For instance, Li \emph{et al.}~\cite{li2017reliable} use crowd-sourcing to simulate human expression recognition, while~\cite{bargal2016emotion, kahou2013combining} employ model ensembling to leverage more information. Farzaneh \emph{et al.}~\cite{farzaneh2021facial} propose a center loss variant to maximize intra-class similarity and inter-class separation for FER, and Ruan \emph{et al.}~\cite{ruan2021feature} acquire expression-relevant information during the decomposition of an expression feature. However, we find that these FER methods are effective when the test set has no domain gap with the train set, while their performance drops drastically when facing domain-different test sets. In this paper, we aim to improve the generalization ability of the FER model and make it suitable for real-world deployment. We train the FER model on one given FER train set and evaluate it on all unseen different FER test sets.

\subsection{Domain Generalization} 
Domain Generalization (DG) aims to help models trained on a set of source domains generalize better on unseen target domains~\cite{ganin2015unsupervised, long2018conditional, yang2022attracting, hur2023learning, dosovitskiy2020image, yang2023tvt, xu2021cdtrans, sun2022safe, rangwani2022closer, kang2019contrastive, long2017deep, tang2020unbiased, lee2019drop}. A common practice is to reduce the feature discrepancy among multiple source domains. \cite{tzeng2014deep, long2015learning, long2017deep} all adopt maximum mean discrepancy on multiple layers to enforce the distribution similarity between source and target features. Deep CORAL~\cite{sun2016deep} uses feature covariance to measure the domain discrepancy. Another stream of works tries to enlarge the available train data space with augmentation of source domains~\cite{carlucci2019domain, dou2019domain, qiao2020learning, shankar2018generalizing, zhou2020deep, zhou2020learning}. Several approaches leverage regularization through domain adversarial learning~\cite{jia2020single, rahman2020correlation} to address DG. However, domain generalization (DG) methods in the FER field usually assume the availability of labeled or unlabeled target expression samples to aid in fine-tuning the FER models. However, our approach differs significantly. We strive to enhance the generalization ability of facial expression recognition (FER) methods by exclusively training the FER model on a single FER dataset and evaluating it on various unseen FER test sets. We refrain from accessing any samples from the target domain for fine-tuning our method, rendering existing FER domain adaptation methods infeasible for our task.

\section{Problem Definition}
In this paper, we aim to improve the generalization ability of the FER model and guide it to recognize unseen test samples with domain gaps of the train samples. Learning is conducted on one given FER train set, and then test samples from different FER test sets with domain gaps of the train set should be recognized on the fly, which is similar to the real-world deployment of FER models.

FER models are trained with $\mathcal{D}_{train}= \left\{ \left( \x _ { i }  , y _ { i }  \right) \right\} _ { i = 1 } ^ {N}$, where $\x_{i}$ is the $i$-th training image and $y_i \in Y  =\{ 1, \ldots , L\}$ is the corresponding label, $N$ is the number of training samples and $L$ is the number of expression classes, we consider the seven basic expressions across different FER datasets in this paper, thus $L$ equals 7. Existing FER models are evaluated on the test set $\mathcal{D}_{test}= \left\{ \left( \x _ { i }  , y _ { i }  \right) \right\} _ { i = 1 } ^ { M }$ that has no domain gap between the train set, $M$ is the number of test samples. However, the real-world test set $\mathcal{D}_{real}= \left\{ \left( \widetilde\x _ { i }  , y _ { i }  \right) \right\} _ { i = 1 } ^ { M }$ is different from $\mathcal{D}_{test}$, as  $\mathcal{D}_{real}$ might contain samples with domain gaps of the training samples. In this paper, we aim to train the FER model on $\mathcal{D}_{train}$ that can generalize well on the real-world test set $\mathcal{D}_{real}$. The difference between our setting and the traditional FER is that we test the FER models on $\mathcal{D}_{real}$ instead of $\mathcal{D}_{test}$. The difference between our setting and domain adaptation FER is that we only use the training set from one source domain and do not have access to the unlabeled samples from $\mathcal{D}_{real}$ to fine-tune the FER model. Our setting is more similar to the real-world deployment as we cannot previously know the distribution of the target domain in the real world.

\section{Method}
\begin{figure}[t]
  \centering
  \includegraphics[width=.98\linewidth]{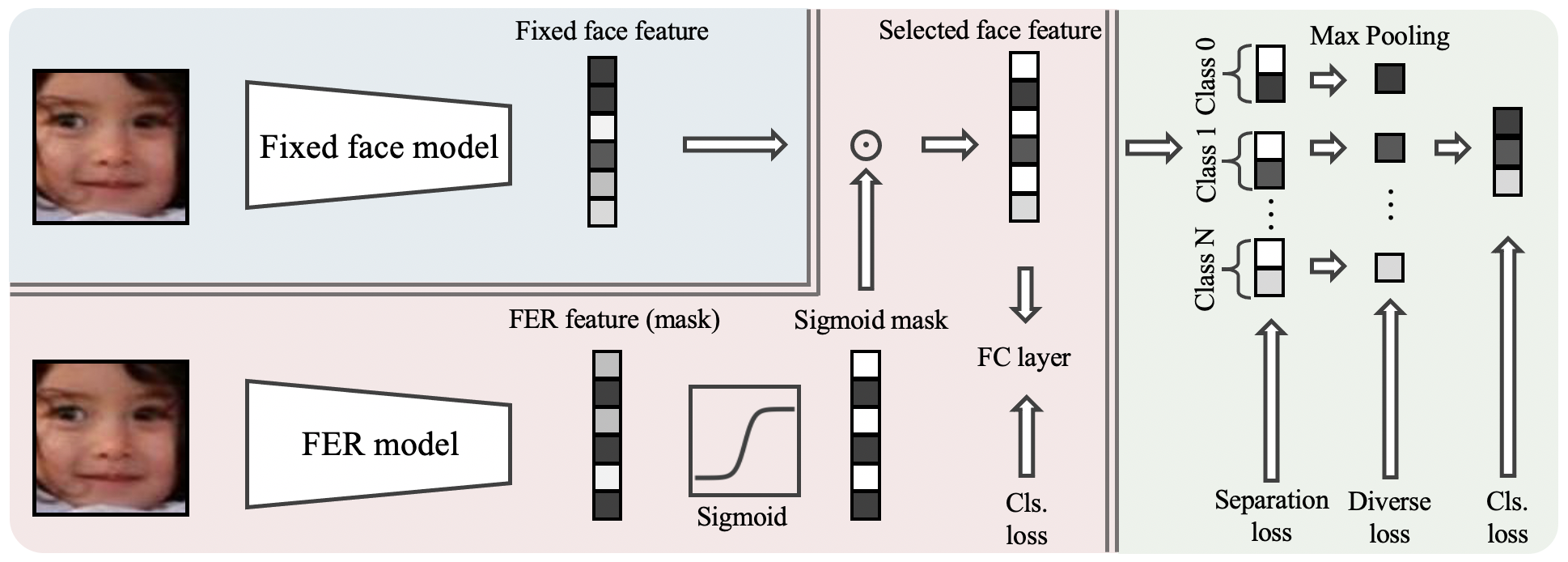}
    \caption{The framework of our proposed method CAFE. We utilize a fixed pre-trained large model, such as CLIP to extract fixed face features regards of the input training images, the FER model is trained to learn a mask for the fixed face features to only extract the expression-related features. Notice that this is similar to how human perceives expressions: we first observe faces and then extract expression-relevant features. The learned mask is regularized by a sigmoid function to prevent overfitting. We further introduce channel-separation and channel-diverse to make the learned mask diverse to improve the generalization ability.}
    \label{pipeline}
\end{figure}

To solve the aforementioned problem, we propose a novel method to mimic the \underline{C}ognition of hum\underline{A}n for \underline{F}acial \underline{E}xpression and name it as CAFE. As humans, when we encounter test samples with domain gaps from the training samples, we first extract their face features to exclude the domain features. Afterward, we focus solely on the features related to the expressions in order to determine the expression contained within the test sample. Following a similar approach, the proposed method initially extracts the face features of the test samples. Then, a trained FER model generates masks for selecting expression-related features from the facial features and making decisions solely based on the selected features. The framework of our proposed method is shown in Fig.~\ref{pipeline}.

\subsection{Mask on Fixed Face Features}
We design a novel method to guide the model to learn masks on fixed face features in order to selectively choose useful expression features. Since face features can be extracted using pre-trained large models like CLIP~\cite{radford2021learning}, we can assume that these features are sufficiently generalizable for different domains. 

Given images $\textbf{x}$ from $\mathcal{D}_{train}$, we first extract the face features using CLIP, denoted as $\textbf{F} \in \mathbb{R}^{N\times C}$, where $N$ is the number of images and $C$ the number of feature dimensions, we fix $\textbf{F}$ during the training process to prevent the FER model to optimize face features to overfit the given train set. This operation improves the generalization ability of our proposed method. The FER model, such as ResNet-18, is trained to learn masks for the given face features. We first extract features $\textbf{f} \in \mathbb{R}^{N \times C \times 1 \times 1}$ from $\textbf{x}$ after the global average pooling (GAP) layer and resize them to generate the masks $\textbf{M} \in \mathbb{R}^{N \times C}$ for face features. Further, in order to regularize the learned masks to generalize to unseen test samples, we apply a sigmoid function on $\textbf{M}$ and get $\textbf{M}_s$ as \begin{equation}
\textbf{M}_s = Sigmoid(\textbf{M}).
\end{equation}
The sigmoid function is vital to the success of our method as it introduces non-linearity into the model, which is crucial for capturing and learning non-linear patterns to generate the masks. The sigmoid function also normalizes $\textbf{M}$, ensuring that the values of $\textbf{M}$ fall within [0, 1], which reduces the overfitting ability of the learned masks. Furthermore, the sigmoid function provides a probability-like output, where the output value represents the probability of selecting the feature of the corresponding channel, which is semantically similar to humans choosing expression-related features from the face features.

We further utilize $\textbf{M}_s$ to select the face features for expression recognition as 

\begin{equation}
\widetilde{\textbf{F}} = \textbf{M}_s \textbf{F}.
\end{equation}

The classification loss is computed between the selected features $\widetilde{\textbf{F}}$ and labels $y$ following 
\begin{equation}
l_{cls} = -\frac1N \sum_{i=1}^N  (\log{\frac{e^{\textbf{W}_{\textbf{y}_i}\widetilde{\textbf{F}}_i}}{\sum\nolimits_{j}^L e^{\textbf{W}_j\widetilde{\textbf{F}}_i}}}), \label{classification}    
\end{equation}
where $\widetilde{\textbf{F}}_i$ is the selected features of image $\textbf{x}_i$, $\textbf{W}_{\textbf{y}_i}$ is the $\textbf{y}_i$-th weight from the FC layer and $\textbf{y}_i$ is the label of $\textbf{x}_i$.

\subsection{Mask Generalization}
To make the learned mask generalizable to other unseen FER test sets, we further design a separation module to regularize the learned mask. Specifically, we set apart the masked features into seven pieces according to the channel dimension to make the masked features correspond to seven basic expressions and then max pool each piece to directly transform them into logits. In such a design, we could avoid the use of the FC layer and directly connect the masked features to the FER labels. The motivation of our design is three folds: Firstly, the learning ability of the FC layer might be too strong to overfit the train set. Thus, we directly transform the masked features into logits to prevent the FC layer from overfitting the labels with the learned features. Secondly, the channel size of 512 when using ResNet-18 might be too large to learn generalizable masks and could also lead to overfitting on the training set. If we set apart the mask into seven pieces and make each piece of the mask correspond to one basic expression, the mask piece with a small channel size will be more likely to only focus on the useful information. Thirdly, setting the masked features into seven pieces is similar to label distribution learning, we guide the seven pieces to focus on features related to the seven basic expressions on one image to solve the FER from a label distribution learning perspective. As FER is similar to a label distribution learning task, one image might contain features of several different expressions. For example, there are compound expressions and one compound expression image contains features from several different basic expression classes.

Specifically, given the masked features $\widetilde{\textbf{F}}$, we divide them according to the channel number $C$ to $L$ pieces corresponding to the class number $L$ on the second dimension, $\widetilde{\textbf{F}} = \{\widetilde{\textbf{F}}_1, \widetilde{\textbf{F}}_2, ..., \widetilde{\textbf{F}}_L\}$. If $C$ can not be divided by $L$, we could divide the selected features non-uniformly. For example, when using ResNet-18, the channel size is 512, we could assign 73 channels for each of the 6 expressions and leave the rest 74 channels for the neural expression. We utilize channel-dropping on the selected features to mitigate the overfitting problem of the selected features. The drop mask is denoted as $\textbf{M}_{drop} = \{\textbf{M}_1, \textbf{M}_2, ..., \textbf{M}_L\}$ with the same size of the selected features. Each mask contains 0 or 1 for feature selection. For example, in the $\textbf{M}_1$, which is a vector of size $(N, 73)$, $N$ is the batch size, we keep the mask same in each batch, thus, for simplicity, we neglect $N$ in the following. On the channel dimension of 73, if we set the drop rate as 10/73, there are 10 channels randomly selected as 0 and all the rest are 1. The channel drop module guides the model to focus on all channels, which increases the generalization ability of the masked features. After channel dropping, the selected features are denoted as \begin{equation}
\overline{\textbf{F}} = \widetilde{\textbf{F}} \textbf{M}_{drop} = \{\widetilde{\textbf{F}}_1 \textbf{M}_1, \widetilde{\textbf{F}}_2 \textbf{M}_2, ..., \widetilde{\textbf{F}}_L \textbf{M}_L\}.
\end{equation} The selected features $\overline{\textbf{F}}$ is downsized to logits $\overline{\textbf{F}}^{d}$ of size $(N,L)$ for classification through a maxpooling operation on the second dimension. 
\begin{equation}
\overline{\textbf{F}}^{d} = \{max(\widetilde{\textbf{F}}_1 \textbf{M}_1), max(\widetilde{\textbf{F}}_2 \textbf{M}_2), ..., max(\widetilde{\textbf{F}}_L \textbf{M}_L)\}.
\end{equation}
Then, we compute a classification loss $l_{sep}$ with the labels and the logits $\overline{\textbf{F}}^{d}$ that are obtained by the separation module without the FC layer. \begin{equation}
l_{sep} = -\frac1N \sum_{i=1}^N  (\log{\frac{e^{ \overline{\textbf{F}}^{d}_{\textbf{y}_i}}}{\sum\nolimits_{j}^L e^{\overline{\textbf{F}}^{d}_j}}}),  
\end{equation}

To increase the generalization ability of the learned masks, we want to make the channels corresponding to each class as diverse as possible. Thus, we further introduce a channel-diverse loss. Specifically, we input the masked features $\widetilde{\textbf{F}} \in \mathbb{R}^{N \times C}$ into the max pooling operation to get 
\begin{equation}
\widetilde{\textbf{F}}_{max} = \{max(\widetilde{\textbf{F}}_1), max(\widetilde{\textbf{F}}_2), ..., max(\widetilde{\textbf{F}}_L)\},
\end{equation} where $\widetilde{\textbf{F}}_{max} \in \mathbb{R}^{N \times L}$. The selected max feature channel of the second dimension of $\widetilde{\textbf{F}}_{max}$ is regularized to be diverse with other feature channels by the channel-diverse loss $l_{div}$,
\begin{equation} l_{div} = 1 - \frac1{Nc} \sum_{i=1}^N \sum_{j=1}^L  \widetilde{\textbf{F}}_{max_{ij}}, \end{equation}
where $i, j$ are utilized to index the first and second dimension of $\widetilde{\textbf{F}}_{max}$, $c$ is the number utilized for normalization. We experimentally set $c$ as 73, which is the same as the channel number of the separated piece for each expression class. The channel-diverse loss regularizes the max value of each piece of $\widetilde{\textbf{F}}_{max}$ as large as possible, which separates the largest value of each piece of $\widetilde{\textbf{F}}_{max}$ from other values, making the values of the channel dimension more diverse.

Combining the separation loss and channel-diverse loss, we aim to learn a powerful while generalizable mask to only select useful expression features from the fixed generalizable face features. The total train loss is summarized as 
\begin{equation}
    l_{train} = l_{cls} + \lambda l_{sep} + \beta l_{div}.
\end{equation}
After training, the module to compute $l_{sep}$ and $l_{div}$ can be abandoned, we only need to keep the module that is used to compute $l_{cls}$ to recognize the test samples as $l_{sep}$ and $l_{div}$ are only used to regularize the $l_{cls}$ during training.

\section{Experiments}

\subsection{Datasets}

RAF-DB~\cite{li2017reliable} is annotated with seven basic expressions by 40 trained human coders, including 12,271 images for training and 3,068 images for testing.

FERPlus~\cite{barsoum2016training} is extended from FER2013 \cite{goodfellow2013challenges} with cleaner labels, which consists of 28,709 training images and 3,589 test images collected by the Google search engine, we utilize the same seven basic expressions with the RAF-DB.

AffectNet~\cite{mollahosseini2017affectnet} is a large-scale FER dataset, which contains eight expressions (seven basic expressions and contempt). There are a total of 286,564 training images and 4,000 test images. We utilize the seven basic expressions in our experiments.

SFEW2.0 is the most commonly used version of SFEW~\cite{dhall2011static}. SFEW2.0 contains 958 train samples, 436 validation samples, and 372 test samples. Each image is assigned to one of the seven basic expressions.

MMA is a large-scale FER dataset with the majority of expressions from individuals of European and American descent. The dataset contains 92,968 training samples, 17,356 validation samples, and 17,356 test samples. Each image is assigned to one of the seven basic expressions.

\begin{table}[!h]
\caption{The test accuracy of FER methods on various FER test sets. The model is only trained using the dataset in the second column (source domain) and evaluated on all five test sets. Our method outperforms SOTA FER methods by large margins on unseen test sets (target domains). We underline the best accuracy of all the compared FER methods and bold the best overall performance. Note that we focus on the generalization ability of FER methods instead of the test accuracy on the source domain.}
\label{main1}
\begin{center}
\footnotesize
\setlength{\tabcolsep}{1mm}
\begin{tabular}{l|c|cccc|c}
\toprule[1pt] 
Method   & RAF-DB & FERPlus & AffectNet &  SFEW2.0 & MMA   & Mean  \\ \midrule

SCN (CVPR2020) &  87.32 & \underline{58.37}   &  42.85    &  44.89  & 36.52 & 53.99 \\
RUL (NeurIPS2021)     &  88.66 & 57.89   &  43.82   &  \underline{46.91}  & 37.11 & 54.88 \\
EAC (ECCV2022)     &  \textbf{89.54} & 54.38   &  \underline{43.91}    &  43.39  & \underline{37.27} & 53.70 \\
OFER (ICCV2023)     &  89.07 & 53.90   & 42.73   &  43.88  & 36.43 & 53.20 \\
CAFE     &  88.72 & \textbf{73.16}  &  \textbf{45.86}    & \textbf{52.86}   & \textbf{56.80}  &  \textbf{63.48} \\ \midrule
Method   & FERPlus & RAF-DB  & AffectNet & SFEW2.0 & MMA   & Mean  \\ \midrule
SCN (CVPR2020)  &  86.80  & \underline{68.71}   &  32.42    &  43.10  & 59.12 & 58.03 \\
RUL (NeurIPS2021) &  88.40  &  51.89  &  35.88    & \underline{45.90}   & 58.00 &  56.01 \\
EAC (ECCV2022) &  89.03  &  58.62  &  \underline{36.49}   & 45.79   & \underline{59.89} & 57.96 \\
OFER (ICCV2023)     &  89.26 & 55.67   &  35.84   &  44.52  & 59.37 & 56.93 \\
CAFE     & \textbf{89.51}  &   \textbf{72.91}  &  \textbf{39.44}    & \textbf{49.38}   & \textbf{60.14}  &  \textbf{62.28} \\ \midrule

Method   & AffectNet & RAF-DB & FERPlus  & SFEW2.0 & MMA   & Mean  \\ \midrule
SCN (CVPR2020) &  62.48    &  \underline{70.70} &  \underline{63.98}   &  41.98  & \underline{38.51} & 55.53\\
RUL (NeurIPS2021) &  58.70    &  55.83 &  52.88   &  34.01  & 31.93  & 46.67 \\
EAC (ECCV2022)  & 64.77 &  66.10 &   57.19  & \underline{44.89}   &  33.49  & 53.09 \\
OFER (ICCV2023)     &  63.90 & 63.21   &  56.79   &  42.60  & 31.99 & 51.70 \\
CAFE     &  \textbf{64.87}    & \textbf{72.69}  & \textbf{69.94}    & \textbf{51.18}   & \textbf{49.65}  & \textbf{60.27}\\ \midrule

Method   & SFEW2.0  & RAF-DB & FERPlus  &  AffectNet  & MMA   & Mean  \\ \midrule
SCN (CVPR2020)  &  51.35   & 46.74  &  32.58    &  23.19  & 17.38 & 34.25 \\
RUL (NeurIPS2021) &  51.87   & 46.35  &  \underline{36.02}    &  \underline{30.41}  & 22.06 & 37.34 \\
EAC (ECCV2022)  &  52.46   &  \underline{47.29} &   33.66  & 25.57 & \underline{22.26}    & 36.25 \\
OFER (ICCV2023)     &  52.15 & 46.33  &  34.78   &  23.02  & 21.88 & 35.63 \\
CAFE &\textbf{53.79}&  \textbf{54.43} &   \textbf{48.39} &\textbf{32.24} & \textbf{36.34}   & \textbf{45.04}\\
\midrule
Method   & MMA       & RAF-DB & FERPlus  & SFEW2.0 & AffectNet   & Mean  \\ \midrule
SCN (CVPR2020)      &  63.00    & 74.09  & \underline{\textbf{73.99}}  &  \underline{45.90}  & \underline{35.94} & 58.58 \\
RUL (NeurIPS2021)     &  61.70    & 71.94  &  69.05  &  39.21  & 34.45 & 55.27 \\
EAC (ECCV2022)      &  65.06    &  \underline{74.32} &   71.85  & 42.87   & 35.83 & 57.99 \\
OFER (ICCV2023)     &  64.74 & 72.85   &  69.69   &  41.49  & 34.58 & 56.67 \\
CAFE &\textbf{65.97}&  \textbf{78.36} &    73.57   & \textbf{49.05}  & \textbf{41.85}  & \textbf{61.76} \\ 
\bottomrule[1pt]
\end{tabular}
\end{center}
\end{table}

\subsection{Implementation Details}
We utilize the most widely employed ResNet-18~\cite{he2016deep} as the backbone following other FER works~\cite{zhang2022learn, wang2020region}. We use the ViT-B/32 CLIP model, only employing its visual component to extract generalizable face features. Any large models trained on extensive face datasets could replace the CLIP model. As studying different models for face feature extraction is not the focus of our paper, we directly utilize the ViT-B/32 CLIP model across all our experiments. The learning rate $\eta$ is set to $0.0002$ and we use Adam~\cite{kingma2014adam} optimizer with weight decay of $0.0001$. We utilize a learning rate scheduler of ExponentialLR, with a gamma of $0.9$. We set the weight for the channel-wise loss as $1.5$ and the weight for the diverse loss as $5$. 


\subsection{Main Experiments of GFER}
To evaluate the generalization ability of existing FER methods, we utilize one of the five FER datasets as the training set. Instead of testing only on the test set of the corresponding train set, we test on all five FER test sets. Note that we \textbf{do not} have access to any labeled or unlabeled images of the target domain. The results are shown in \cref{main1}. Our comparison methods are FER methods published in the top conferences in recent years, such as SCN~\cite{wang2020suppressing}, RUL~\cite{zhang2021relative}, EAC~\cite{zhang2022learn}, and OFER~\cite{lee2023latent}.

\noindent\textbf{Generalization Ability} The results reveal that SOTA methods in facial expression recognition (FER) do not exhibit strong performance when applied to FER test datasets characterized by domain gaps relative to the training set. For instance, while an advanced FER method like EAC significantly enhances performance over the SCN method on the test set that corresponds to the training set, its results on other FER test datasets are comparable to, or even worse than, those of the SCN. We speculate the reason might lie in that EAC fits the train set too well, which improves the test accuracy on the corresponding test set and degrades the test accuracy on other different FER test sets. We underline the best result of other FER methods and compare it with our method. Our method outperforms other FER methods by large margins under almost all settings. Under different FER train sets, our method always achieves the best mean accuracy on five different FER test sets. We also show the test accuracy of each expression class and the mean accuracy in the Supp. material and find that our method also achieves the best mean test accuracy of different expression classes.

\noindent\textbf{Cross-Domain Analysis} We observe that test accuracy on RAF-DB and FERPlus are the highest across other datasets. Using SFEW2.0 as the train set achieves the lowest test accuracy on other test sets, as SFEW2.0 is the smallest dataset used in our paper and some images have low quality. Test accuracy on AffectNet is always low given different train sets because the labels of AffectNet are very noisy. Some error examples are shown in the Supp. material to help understanding.

\subsection{Discussion} 
\textbf{Our paper focuses on the generalization ability of FER methods instead of the test accuracy on the source domain.} Actually, we find that strong FER methods like EAC that perform well on the test set of source domain only fits the train set well and cannot generalize to other different unseen test sets. Thus, in this paper, we aim to propose a method that can perform well on the source domain test set and generalize to other unseen test sets at the same time. We do not claim that our method always achieves the best performance on the test set of the source domain, which is not our main focus. Based on the experimental results in \cref{main1}, we argue that our method performs on par with SOTA FER methods on different source domain test sets and achieves the best performance on different unseen test sets, which completely fulfills our goal.

\subsection{Ablation Study}
To study the effectiveness of each of the proposed modules in our method, we carry out thorough ablation studies. The results in \cref{ablation} show that without our method, the FER model cannot generalize to other datasets that have domain gaps with the train set, which is unsatisfactory for the real-world deployment of FER models. With our proposed sigmoid mask learning, the performance on other unseen test sets outperforms the baseline by a large margin. However, the fitting probability of the mask is too strong as the dimension of 512 might be too much to learn generalizable expression features. Thus, we further introduce the separation module which separates the masked features into pieces corresponding to the number of expression classes. It shows that the results improve from only using the sigmoid mask module. We also introduce a channel-diverse loss to make the channels in each piece as diverse as possible, which further improves the accuracy based on using the sigmoid mask and the channel-separation module. From the results in \cref{ablation}, each module contributes to the success of our proposed method, and combining them together achieves the best result.

\begin{table}[t]
\begin{center}
\caption{The ablation study of our proposed method. The results on different FER test sets show that the most effective module of our method is the mask module. The separation and diverse module make the learned mask more generalizable to other FER test sets and further improve the performance of our method.}
\label{ablation}
\setlength{\tabcolsep}{2mm}
\begin{tabular}{ccc|cccc|c}
\toprule[1pt]
Mask & Separation & Diverse & FERPlus  & AffectNet &  SFEW2.0         & MMA      & Mean \\ \midrule[1pt]
     &            &         &  58.05   &  43.25    &  42.76           &  42.61   & 46.67    \\
 \checkmark    &  &         &  70.90   &  43.77    &  51.63           &  55.65   & 55.49   \\
 \checkmark & \checkmark &  &  72.01   &  45.17    &  \textbf{53.31}  &  56.69   & 56.80    \\
 \checkmark & \checkmark &\checkmark& \textbf{73.16} & \textbf{45.86} &  52.86   &  \textbf{56.80}    & \textbf{57.17}    \\ 
 \bottomrule[1pt]
\end{tabular}
\end{center}
\end{table}

\subsection{Different Backbones}
We combine our method with different backbones to show its generalization ability. The results in \cref{backbone} illustrate that our method improves the performance of baseline under different backbones by remarkable margins. Specifically, the improvement is largest when using ResNet-18 as the backbone, the reason might lie in that the dimension of the output feature of ResNet-18 is 512, which is the same as the dimension of the output feature of CLIP. When using MobileNet~\cite{howard2017mobilenets} or ResNet-50, we reduce the size of the output feature through mean operation to suit the feature dimension of CLIP, which might slightly limit the performance improvement. For example, when the backbone is ResNet-50, we simply reduce the output feature dimension of 2048 to 512 through mean operation using sliding windows. We also observe that stronger backbones have better generalization ability in our experiment. The ResNet-50 backbone achieves the overall best performance across experiments.

\subsection{Comparison with CLIP+Finetune}
To demonstrate the significant generalization ability inherent in our proposed method, we construct a comparison group comprising a fixed CLIP model alongside a trainable FER model, which has similar trainable parameters with our method for fair comparison. We then fine-tune a fully connected layer on the concatenated features. The results of this comparison are depicted in \cref{CLIP_compa}, confirming that it is the specific design of our method that effectively unleashes CLIP's generalization potential for FER.

\begin{table}[t]
\caption{The performance of our method under different backbones. Our method could improve the performance on unseen FER test sets under different backbones.}
\label{backbone}
\begin{center}
\setlength{\tabcolsep}{1.5mm}
\begin{tabular}{l|c|cccc|c}
\toprule[1pt]
Backbone      & RAF-DB     & FERPlus  & AffectNet    & SFEW2.0    & MMA    & Mean  \\ \midrule[1pt]
MobileNet     & 84.65      & 61.33    &   \textbf{43.45}      & 43.21      & 40.67  & 54.66 \\
MobileNet + CAFE &  \textbf{85.07}  &  \textbf{64.97}   &   42.91      & \textbf{45.45}      & \textbf{43.11}  &\textbf{56.30} \\   \midrule
ResNet-18      &   88.40   & 58.05    &  43.25       & 42.76      & 42.61  & 55.01 \\
ResNet-18 + CAFE & \textbf{88.72}   & \textbf{73.16}    &  \textbf{45.86}       & \textbf{52.86}      & \textbf{56.80}  & \textbf{63.48} \\ \midrule
ResNet-50    & 88.49       &   70.13  & 47.46        & 49.94      & 48.90  & 60.98 \\
ResNet-50 + CAFE & \textbf{89.05}   & \textbf{75.26}    & \textbf{47.49}        & \textbf{51.07}      & \textbf{57.46}  & \textbf{64.07} \\ 
\bottomrule[1pt]
\end{tabular}
\end{center}
\end{table}

\begin{table}[t]
\caption{Comparison with CLIP+Finetune. The comparison group has the similar trainable parameters with our method for fair comparison. Our method outperforms the comparison group under all settings, illustrating that it is our designed method that unleashes the generalization ability of CLIP for FER.}
\label{CLIP_compa}

\begin{center}
\footnotesize
\setlength{\tabcolsep}{2mm}
\begin{tabular}{l|c|cccc|c}
\toprule[1pt] 
Method   & RAF-DB & FERPlus & AffectNet &  SFEW2.0 & MMA   & Mean  \\ \midrule[1pt]
CLIP+Finetune& 88.69 & 59.71 & 44.31 & 41.30 & 44.62 & 55.73\\ \midrule
\multirow{2}{*}{CAFE}  &  \textbf{88.72} & \textbf{73.16}  &  \textbf{45.86}& \textbf{52.86}& \textbf{56.80} &  \textbf{63.48} \\ 
& \textcolor{blue}{+0.03} & \textcolor{blue}{+13.45} &  \textcolor{blue}{+1.55}    & \textcolor{blue}{+11.56}   &  \textcolor{blue}{+12.18} &  \textcolor{blue}{+7.75} \\ 
\bottomrule[1pt]
\end{tabular}
\end{center}
\end{table}

\begin{table}[t]
\caption{Influence of the sigmoid function on our method. Our method uses the sigmoid mask to select expression-related features, the comparison group has no sigmoid function and the others are the same as our method. The results show that the sigmoid function is very important for our method to learn generalizable masks.}
\label{sigmoidtable}
\begin{center}
\setlength{\tabcolsep}{2mm}
\begin{tabular}{l|c|cccc|c}
\toprule[1pt] 
Method   & RAF-DB & FERPlus & AffectNet &  SFEW2.0 & MMA   & Mean  \\ \midrule[1pt]
No sigmoid   &  \textbf{89.24} & 62.58   &  \textbf{46.37}    &  46.58 & 45.49 & 58.05 \\ \midrule
\multirow{2}{*}{CAFE}      &  88.72 & \textbf{73.16}  &  45.86   & \textbf{52.86}    & \textbf{56.80}  &  \textbf{63.48} \\ 
    & \textcolor{red}{-0.52} & \textcolor{blue}{+10.58} &  \textcolor{red}{-0.51}    & \textcolor{blue}{+6.28}   &  \textcolor{blue}{+11.31} &  \textcolor{blue}{+5.43} \\ 
\bottomrule[1pt]
\end{tabular}
\vspace{-1mm}
\end{center}
\end{table}

\begin{figure}[]
  \centering
  \includegraphics[width=1.0\linewidth]{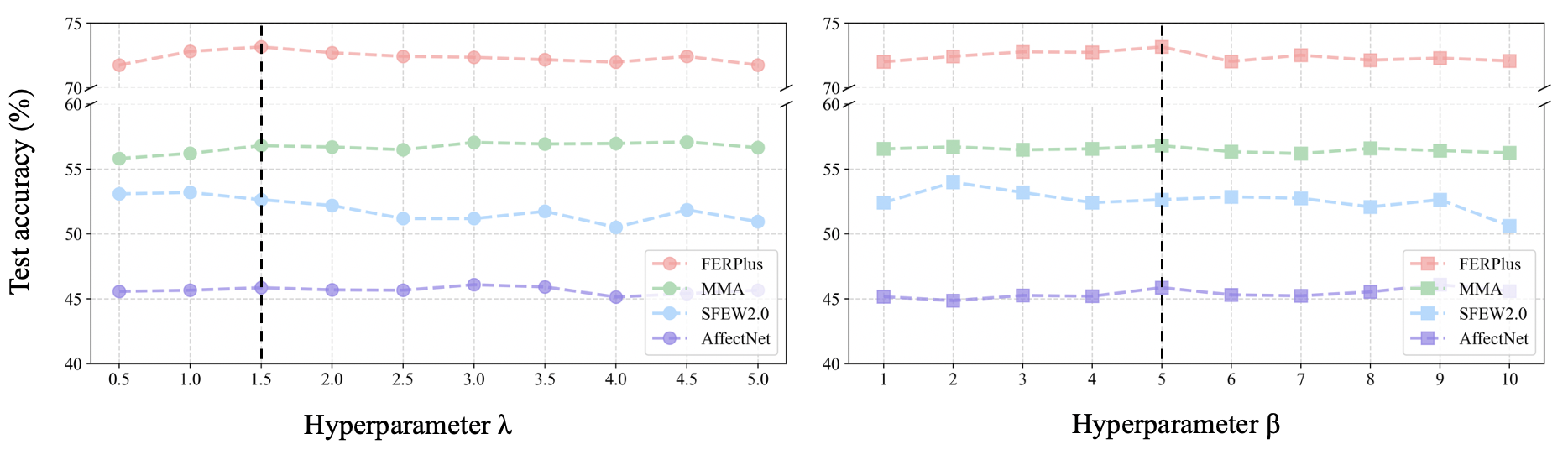}
    \caption{The hyperparameter study of our method. Our method is not very sensitive to the two hyperparameters and we could choose them from a wide range. For simplicity, we use $\lambda = 1.5$ and $\beta = 5$ across all experiments. The cases when $\lambda$ is $0$ or when $\beta$ is $0$ are studied in \cref{ablation}.}
    \label{hyperparameter}
      \vspace{-4mm}
\end{figure}

\subsection{The Effect of the Sigmoid Function}
To illustrate the effectiveness of the sigmoid function applied to the learned mask, we design a comparison group of our method without the sigmoid function, while we keep the others exactly the same. The experiment results on different FER test sets are shown in \cref{sigmoidtable}. The results demonstrate that the sigmoid function for mask learning is very important, without it, our method barely works. The reason lies in that the sigmoid function normalizes the learned masks, ensuring that the values fall within [0, 1], which reduces the overfitting ability of the learned masks.

\subsection{Hyperparameter Study}
We study the influence of the weight of separation loss $\lambda$ and the weight of diverse loss $\beta$ on our method. The results shown in Fig.~\ref{hyperparameter} illustrate that both the $\lambda$ and $\beta$ can be chosen from a wide range and the performance is only slightly different across an order of magnitude, e.g., $\lambda$ from $[0.5, 5]$ and $\beta$ from $[1, 10]$. For simplicity, we choose $\lambda$ as $1.5$ and $\beta$ as $5$ across all our experiments. The cases when $\lambda$ is $0$ or when $\beta$ is $0$ are studied in \cref{ablation}.

\subsection{Visualization Results}

\begin{figure}[t]
  \centering
  \includegraphics[width=1.\linewidth]{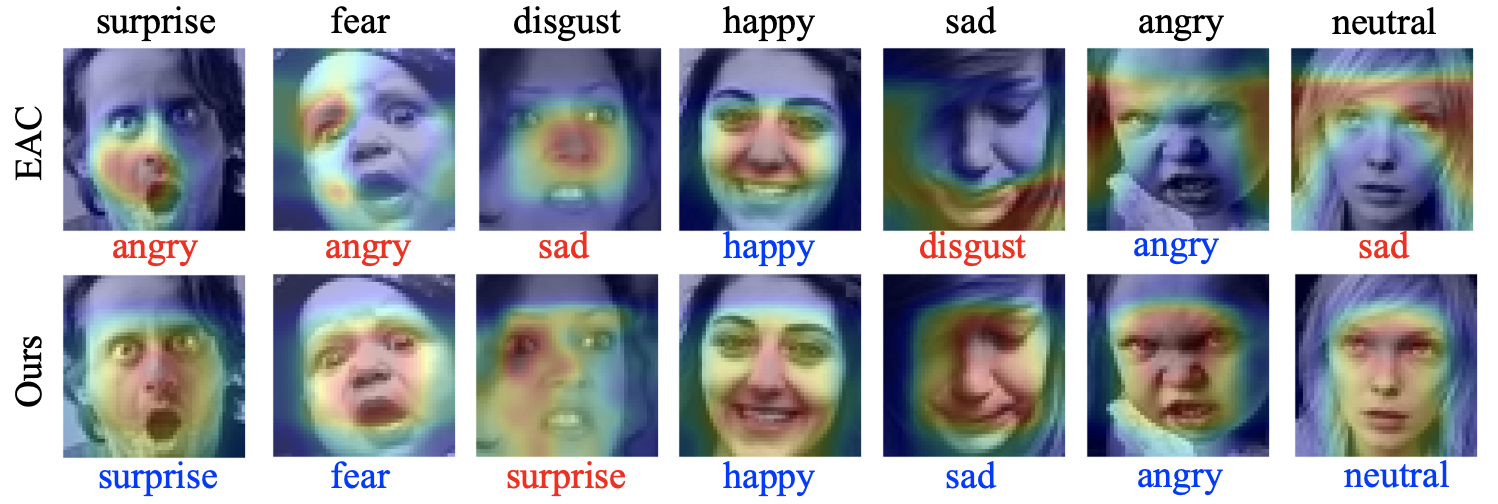}
   \caption{The extracted features of FERPlus test samples by the RAF-DB trained model of EAC and our method. Labels are displayed in black, correct predictions are displayed in blue, and incorrect predictions in red.}
    \label{cross}
\end{figure}

\begin{figure}[!t]
  \centering
  \includegraphics[width=1.\linewidth]{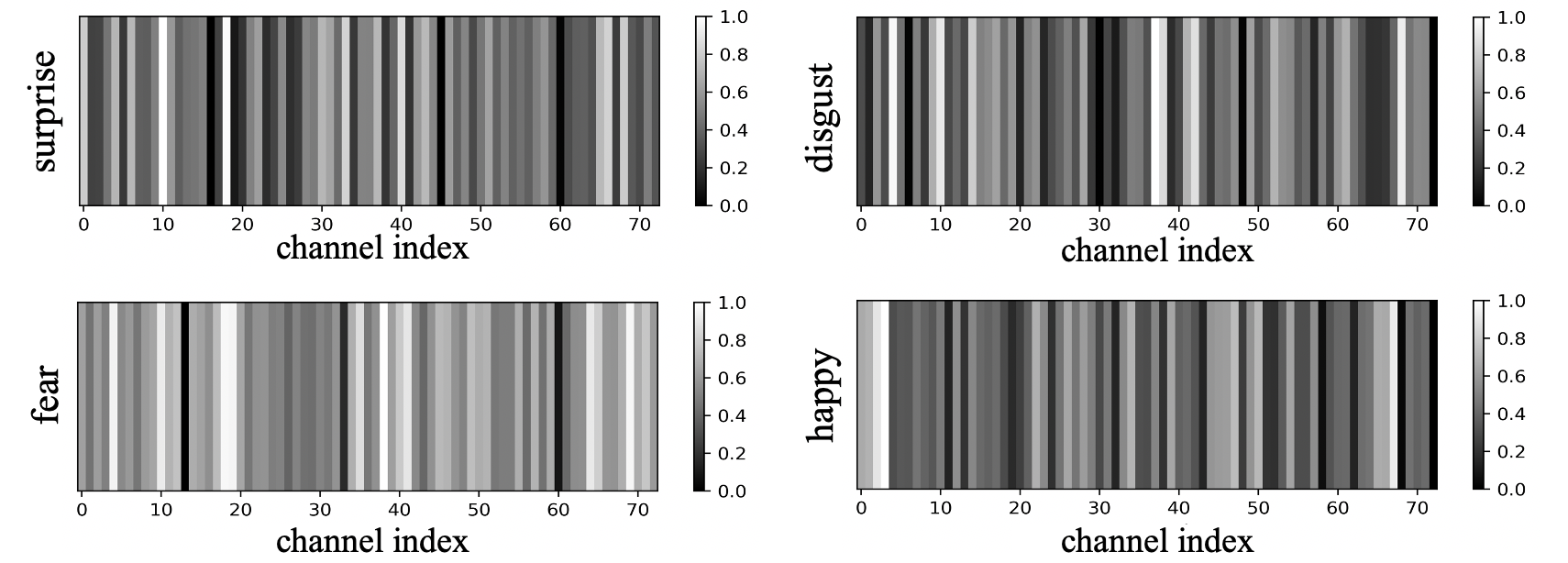}
   \caption{Visualization of the sigmoid masks.}
    \label{binary}
    \vspace{-4mm}
\end{figure}

We use GradCAM~\cite{selvaraju2017grad} to visualize the features of FERPlus test samples by the models trained on RAF-DB. Results are shown in Fig.~\ref{cross}. As both EAC and our method utilize ResNet-18 instead of Transformers as the backbone, the learned features do not lie in some local areas. According to our visualization results and analyses in paper~\cite{zhang2022learn}, EAC focuses on wrong partial features, while our method focuses on the whole face, learns better expression-related features on unseen test samples. We also plot the learned sigmoid masks in Fig.~\ref{binary}, the learned masks corresponding to different expression classes show clearly different patterns, which implicitly illustrate the effectiveness of our method as some feature channels from the fixed face model are constantly useful for specific expression classes.

\section{Conclusion}
In this paper, we aim to improve the zero-shot generalization ability of FER methods on different unseen test sets with only one train set. Enlightened by how humans first detect faces and then recognize expressions, we propose a novel method to learn sigmoid masks on fixed general face features to extract expression-related features, combining the generalization ability of large models extracted features and the high precision of FER models. To further improve the generalization ability on unseen test sets, we propose channel-separation and channel-diverse modules to regularize the learned sigmoid masks. Extensive experiments on five different FER datasets illustrate that our method achieves the best generalization ability across SOTA FER methods.

\section*{Acknowledgements}
We sincerely thank the ACs of this paper. This work was supported by the National Natural Science Foundation of China under Grant No.62276030. 

%
%
\bibliographystyle{splncs04}
\bibliography{main.bbl}
\end{document}